\documentclass[10pt,journal,compsoc]{IEEEtran}
\usepackage{amsmath,amsfonts}
\usepackage{algorithmic}
\usepackage{algorithm}
\usepackage{array}
\usepackage[caption=false,font=normalsize,labelfont=sf,textfont=sf]{subfig}
\usepackage{textcomp}
\usepackage{stfloats}
\usepackage{url}
\usepackage{verbatim}
\usepackage{graphicx}
\usepackage{booktabs}
\usepackage{cite}
\usepackage{soul}
\usepackage{xcolor,colortbl}
\usepackage[font=small,labelfont=bf,tableposition=top]{caption}
\usepackage[colorlinks=true]{hyperref}
\usepackage{threeparttable/threeparttable}
\usepackage{xcolor}
\newcommand{\methodname}{{\fontfamily{ppl}\selectfont
Wild2Avatar}}
\definecolor{Highlight}{HTML}{000000}  
\definecolor{R02}{HTML}{000000}  

\newcolumntype{P}[1]{>{\centering\arraybackslash}p{#1}}
\newlength\savewidth

\definecolor{best_color}{HTML}{FCE5CD}
\definecolor{human_color}{HTML}{ED7D31}
\definecolor{obstacle_color}{HTML}{4472c4}
\definecolor{background_color}{HTML}{70AD47}

\begin{document}

\title{Rendering Humans Behind Occlusions}
\author{Tiange Xiang, 
        Adam Sun, 
        Scott Delp,
        Kazuki Kozuka,
        Li Fei-Fei*,
        Ehsan Adeli*
\IEEEcompsocitemizethanks{
\IEEEcompsocthanksitem * Equal mentorship; Correspondence to: \{xtiange,eadeli\}@stanford.edu.
}
}

\markboth{IEEE TRANSACTIONS ON PATTERN ANALYSIS AND MACHINE INTELLIGENCE, 2025}%
{Shell \MakeLowercase{\textit{et al.}}: Bare Demo of IEEEtran.cls for Computer Society Journals}

\IEEEtitleabstractindextext{%
\begin{abstract}
Rendering the visual appearance of moving humans from occluded monocular videos is a challenging task. Most existing research renders 3D humans under ideal conditions, requiring a clear and unobstructed scene. Those previous methods cannot be used to render humans in real-world scenes where obstacles may block the camera's view and lead to partial occlusions. In this work, we present \textbf{\methodname}, a neural rendering approach catered for occluded in-the-wild monocular videos. We propose occlusion-aware scene parameterization for decoupling the scene into three parts - occlusion, human, and background. Additionally, extensive objective functions are designed to help enforce the decoupling of the human from both the occlusion and the background and to ensure the completeness of the human model. Wild2Avatar is verified with experiments on 14 challenging in-the-wild videos.
\end{abstract}

}  

    


\maketitle

\section{Introduction}\label{sec:intro}
Rendering humans from videos has a wide variety of applications in many fields, including augmented/virtual reality \cite{guo2023vid2avatar, habermann2023hdhumans}, film \cite{carranza2003free}, and healthcare \cite{gerats2022depth}. Videos from single cameras are widespread and easy to acquire, so rendering humans from monocular videos has been the target of copious research. 
Methods to accomplish this task such as Vid2Avatar \cite{guo2023vid2avatar}, MonoHuman \cite{yu2023monohuman}, and NeuMan \cite{jiang2022neuman} have achieved impressive performance. However, most existing human rendering methods are designed for ideal experimental scenes with little to no obstacles and a full view of the human in each frame. In real-world scenes, however, humans might move behind objects and move them around, causing the objects to block the camera's view.

\textcolor{R02}{Most neural rendering methods face significant challenges when dealing with occlusions in real-world scenarios, primarily due to insufficient supervision. Unlike controlled environments, in-the-wild settings typically lack ground-truth reference for human appearance, geometry, and pose. As a result, the model must infer these attributes from limited and often partial visual evidence. This inference becomes especially difficult when large portions of the human body are occluded. Additionally, many existing methods rely on point-based rendering schemes, which can produce vastly different outputs when adjacent points are occluded \cite{liu2022neuralray, xiang2023rendering}. Methods that are not specifically adapted to handle occlusions tend to generate incomplete human bodies with visual artifacts and floaters. This problem is particularly critical in surveillance videos from environments such as construction sites and hospitals, where humans are frequently blocked by various objects. Accurate reconstruction and rendering of occluded humans in these contexts is therefore essential, as it directly affects performance in downstream tasks.}

\begin{figure}[t]
  \centering
  \includegraphics[width=1.0\linewidth]{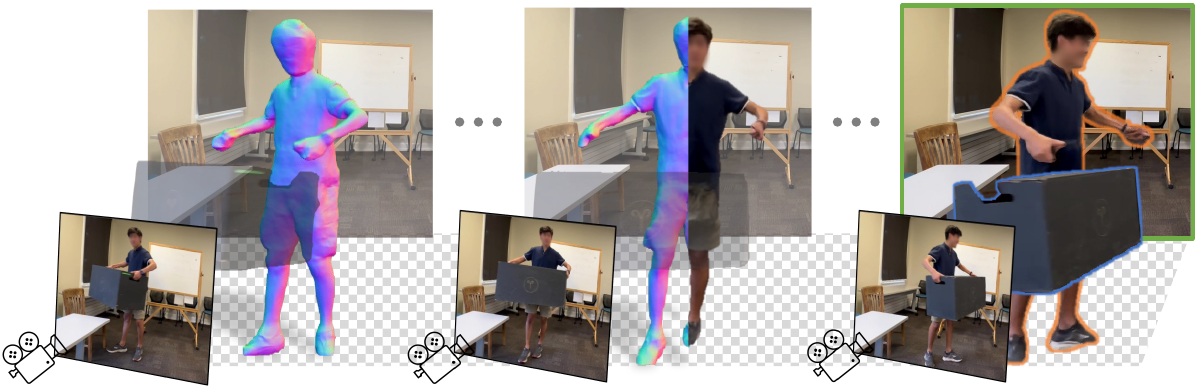}
  \caption{In this paper, we introduce \textbf{\methodname}, a method to render high fidelity human avatars from in-the-wild monocular videos behind occlusions. To achieve photo-realistic rendering, we parameterize the scene into three parts: \textcolor{obstacle_color}{occlusion}$\rightarrow$\textcolor{human_color}{human}$\rightarrow$\textcolor{background_color}{background} and decouple the renderings through novel optimization objectives. Faces in the figure are blurred for anonymity.}
  \label{fig:fig1}
\end{figure}

In this work, we address these shortcomings by introducing \methodname ~(\figureautorefname~\ref{fig:fig1}), an improved method for rendering higher-fidelity humans under occlusion based on our previous ICCV-2023 publication OccNeRF \cite{xiang2023rendering}. 

Our method builds off a novel scene parameterization technique, occlusion-aware scene parametrization that divides the whole scene into three parts --- occlusions, human, and background. By extending upon inverted sphere parametrization \cite{zhang2020nerf++} with a second, inner sphere, and by defining the region from the camera to the edge of the inner sphere as the occlusion region, our method allows for a clean 3D reconstruction of the human regardless of occlusion.  To ensure a high-fidelity and complete rendering of humans, we propose aggregating the three components through a combination of a pixel-wise photometric loss, a scene decomposition loss, an occlusion decoupling loss, and a geometry completeness loss. \emph{Along with OccNeRF \cite{xiang2023rendering}, our work is the first neural method for cleanly rendering humans from occluded monocular videos free of floaters.} Similar to most previous literature in human rendering \cite{weng2022humannerf,guo2023vid2avatar,xiang2023rendering}, our method mainly focuses on reconstructing the human – we do not aim to accurately render irrelevant occlusions or the background in this work.




In summary, our contributions in this paper are:
\begin{enumerate}
    \item We introduce occlusion-aware scene parametrization, a method to decouple a scene into three parts: occlusion, human body, and background. 
    \item We propose a new rendering framework that renders each of these three parts separately and design novel optimization objectives to ensure a clean decoupling of the occlusion and a complete human rendering. 
    \item We evaluate our method on 14 challenging occlusion-intensive in-the-wild videos, demonstrating its effectiveness in rendering occluded humans.
\end{enumerate}

\section{Related Work}
\label{sec:relatedwork}
\noindent
\subsection{3D Human Modeling and Rendering}
There have been numerous works focused on free-viewpoint rendering of humans. While past works were able to achieve good quality renderings of humans from dense \cite{relightables, debevec, zhou2023relightable} and sparse \cite{peng2021neural, zhou2023hdhuman, deepvolumetric, xu2021hnerf, liu2022neural, li2022tava, peng2021animatable} camera views, a recent research focus involves rendering a moving human from a single camera angle \cite{weng2022humannerf, yu2023monohuman, guo2023vid2avatar, jiang2022selfrecon, jiang2022instantavatar, alldieck2018detailed, alldieck2018video, jiang2022hifecap, habermann2019livecap, jayasundara2023flexnerf}. To accomplish this task, Neural Radiance Fields (NeRFs) \cite{nerf} based methods have been more favored recently due to their high rendering quality. Originally used to model static scenes, NeRFs have been adapted to model a dynamic human by parameterizing the human body using the SMPL \cite{SMPL} prior and mapping from the pose-independent canonical space to the observation space \cite{weng2022humannerf, yu2023monohuman, guo2023vid2avatar, jayasundara2023flexnerf}.  HumanNeRF was one of the first to work on free-viewpoint rendering of humans in complex motion from monocular video \cite{weng2022humannerf}. MonoHuman models the deformation field with bi-directional constraints for better rendering of humans \cite{yu2023monohuman}. 

More recent works in neural rendering harness 3D Gaussian splatting \cite{kerbl20233d}, which allows for extremely efficient training and rendering of scenes. By initializing the Gaussians based on the SMPL prior, numerous concurrent works have been able to render humans extremely efficiently and accurately \cite{hu2023gauhuman, kocabas2023hugs, moreau2023human, qian20233dgs, li2023human101}.

However, while these existing methods perform well in a clean environment, they mostly fail when occlusions block the camera's view of the human.

\vspace{5pt}
\noindent
\subsection{Rendering via Scene Decomposition}
Complex natural scenes are usually composed of multiple sub-components. So, a common approach to render these scenes is to model the different 3D components of the scene individually and then aggregate them together in the 2D image space \cite{wu2022d2nerf, jiang2022neuman, niemeyer2021giraffe, guo2023vid2avatar, yuan2020star}. 
GIRAFFE \cite{niemeyer2021giraffe} was one of the pioneering works that proposed using generative neural fields to disentangle objects from the background. The same idea can be also found in NeRF++ \cite{zhang2020nerf++}, which is specifically designed for rendering unbounded surroundings, with two neural fields being trained separately for foreground and background. STNeRF \cite{zhang2021stnerf} utilizes neural layers to render humans and background separately from multiple cameras, while NeuMan \cite{jiang2022neuman} trains a separate human and scene radiance field on a monocular in-the-wild video. In \sectionautorefname~\ref{sec:evaluations}, we detail why NeuMan is not able to be used as a major comparison baseline. Vid2Avatar uses inverse sphere parametrization from \cite{zhang2020nerf++} to separate the dynamic human from the static background, resulting in clean and detailed human avatars \cite{guo2023vid2avatar}. Vid2Avatar's high-fidelity yet robust renderings with minimal artifacts inspire us to use it as the foundation of this work. \textcolor{Highlight}{More recently, HSR \cite{xue2024hsr} introduced a unified framework that jointly reconstructs dynamic humans and static scenes in a shared global coordinate system from monocular RGB videos. Unlike prior works that process humans and scenes separately, HSR explicitly models physical interactions, such as occlusions and interpenetration, enabling more coherent and realistic reconstructions.}
\noindent
\subsection{Occlusion Handling}
While accounting for occlusions in rendering has been a long-standing research problem \cite{Xue2015, li2020recurrent, nazeri2019edgeconnect, jonna2017stereo,mu2013video}, occlusion handling for NeRFs is relatively new. OCC-NeRF \cite{Zhu2023} uses a scene MLP and a background MLP to remove occlusions from the output. NeuRay \cite{neuray} utilizes feature vectors to track the visibility of each 3D point to determine whether it is occluded. While these approaches reduce artifacts caused by occlusion, they require multiple input views to predict visibility and are thus not generalizable to an in-the-wild monocular video of a dynamic human.
HOSNeRF \cite{liu2023hosnerf} can render human, objects, and the scene separately from a monocular in-the-wild video by introducing object bones that are attached to the human skeleton. However, this approach is only robust to very small-scale occlusions and is only applicable for dynamic scenes due to the constraints of COLMAP. We detail why HOSNeRF is not able to be used as a major comparison baseline in \sectionautorefname~\ref{sec:evaluations}.

\subsection{Connection to Our Previous Work}
We first introduced the problem of occluded human rendering from monocular videos along with a solution in our previous ICCV-2023 publication OccNeRF \cite{xiang2023rendering}. This paper is a significant extension with the following three major improvements:
\begin{enumerate}
    \item We have thoroughly revised the OccNeRF pipeline and present novel methodologies for both scene modeling and human rendering. 
    \item We have evaluated the proposed methods on seven more challenging in-the-wild videos with real-world occlusions. Extensive experiments were conducted for further study on the robustness towards the number of training images as well as the model's generative ability.
    \item We achieve superior results that renders high-fidelity human bodies with much less artifacts and floaters.
\end{enumerate}
\begin{figure*}[t]
    \centering
    \includegraphics[width=0.99\linewidth]{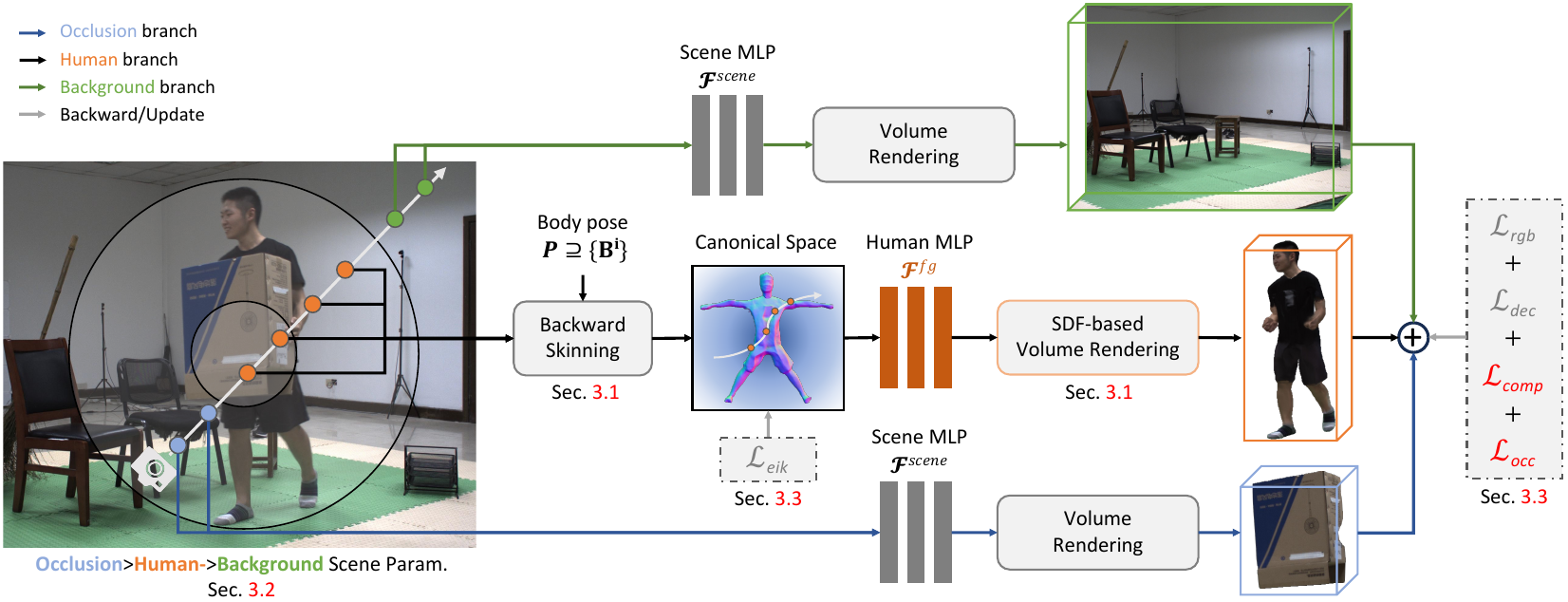}
    \caption{\textbf{\methodname~ renders occluded humans from a monocular video.} It parameterizes the 3D scene into three parts: \textcolor{obstacle_color}{occlusion}, \textcolor{human_color}{human}, and \textcolor{background_color}{background}, in order from closest to farthest from the camera. The human and the occlusions/background are individually modeled via separate neural radiance fields. The \textcolor{human_color}{human}  \textcolor{Highlight}{is parameterized in a bounded space between an inner sphere $\Pi$ with radius $\mathbf{r}$ and an outer sphere with radius $\mathbf{R}$} by first deforming ray samples into a canonical space with the help of the pre-computed body pose $P$. The canonical points $\mathbf{x}$ are passed into a rendering network $\mathcal{F}^{fg}$ to learn the radiance $\mathbf{c}$ and distance $\mathbf{s}$ to the surface of the human, which can then be rendered through SDF-based volume rendering. The unbounded \textcolor{background_color}{background} \textcolor{Highlight}{samples outside the outer sphere are represented as coordinates on the surface of the outer sphere along with their inverted distances.} Another rendering network $\mathcal{F}^{scene}$ is used to learn the radiance and density for the background ray samples \textcolor{Highlight}{(shown as green dots )}. The space of \textcolor{obstacle_color}{occlusion} is determined as the interval between the camera and the inner sphere $\pi$. We parameterize ray samples as coordinates on the surface of $\pi$ and the negation of the inverted distances to the center of the inner sphere. We rely on the same network $\mathcal{F}^{scene}$ to render the occlusions. The three renderings are sequentially aggregated and supervised on a combination of losses, in which we specifically encourage the decoupling of the occlusion from the human through $\mathcal{L}_{occ}$ and penalize the incompleteness of human geometry through $\mathcal{L}_{comp}$.}
    \label{fig:framework}
\end{figure*}

\section{Methods}
\label{sec:methods}

We present \methodname, a model that renders 3D humans with complete geometry and high-fidelity appearance for in-the-wild monocular videos with occlusions. We start by reviewing the basics of NeRFs and the kinetics of the human body (\sectionautorefname~\ref{sec:background}). We then present the key concepts of scene parameterization, which is central to the success of occlusion-aware human renderings (\sectionautorefname~\ref{sec:scene_param}). To follow, we outline the objective functions needed for complete human modeling and crisp scene decomposition (\sectionautorefname~\ref{sec:objectives}). Lastly, we tie all the components together and outline the overall framework of \methodname\ in \figureautorefname~\ref{fig:framework}. 

\subsection{Preliminaries and Background} \label{sec:background}

\noindent
\textbf{Implicit neural radiance fields.} NeRFs~\cite{nerf} learn a neural network $\mathcal{F}$ to model the mappings between (positionally embedded) 3D coordinates $\{\mathbf{x}\in\mathbb{R}^{3}\}$ and their radiance $\mathbf{c}(\mathbf{x})$ and density $\sigma(\mathbf{x})$. This representation is applicable to stationary one-object scenes but usually suffers from unexpected rendering artifacts and floaters in the wild. For rendering solid and continuous human bodies, using an implicit surface representation with a Signed Distance Function (SDF) is preferred \cite{park2019deepsdf, yariv2021volume, wang2021neus, guo2023vid2avatar}. Instead of learning density values for each $\mathbf{x}$, $\mathcal{F}$ is trained to output distances to the surface of the human body $\mathbf{s}(\mathbf{x})$. The human geometry is then represented as a surface model by the zero-level set: $\{\mathbf{s}(\mathbf{x})=0\}$.

\noindent
\textbf{Human body deformation.} Following \cite{weng2022humannerf, chen2021snarf}, we model an articulated human in a static canonical space $\{\mathbf{x}_c\}$ and deform its body from and to the observation space $\{\mathbf{x}_o\}$ via backward and forward skinning~\cite{chen2021snarf, guo2023vid2avatar}:
\begin{equation}
    \mathbf{x}_o = \sum_{i}w^i_c\mathbf{B}^i\mathbf{x}_c,\ \ \ \ \ \ \ \ \mathbf{x}_c = (\sum_{i}w^i_o\mathbf{B}^i)^{-1}\mathbf{x}_o,
\end{equation}
where $\mathbf{B}$ is the bone transformation and $\{w^i_{(\cdot)}\}$ are vertex-wise skinning weights derived from SMPL \cite{SMPL} poses $\mathbf{P}$. We are then able to first optimize a static neural field for the dynamic human in the canonical space and then deform the ray samples to the observation space for volume rendering \cite{park2021nerfies,pumarola2021d,peng2021animatable}.

\subsection{Occlusion-aware Scene Parameterization} \label{sec:scene_param}

\begin{figure}[t]
    \centering
    \includegraphics[width=0.95\linewidth]{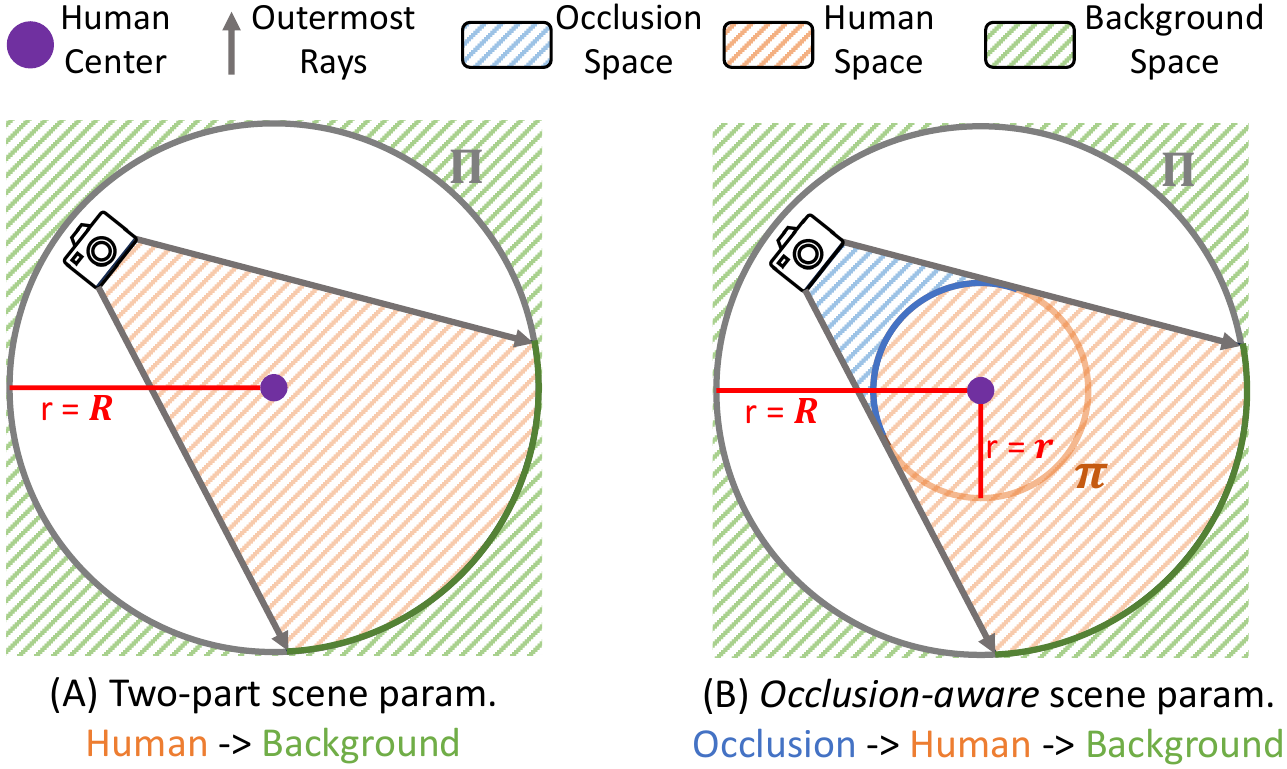}
    \caption{\textbf{Different Scene parameterizations in 2D simplifications.} \textbf{(A)} The regular two-part paradigm \cite{zhang2020nerf++} parameterizes the foreground within a sphere $\Pi$ with radius $\mathbf{R}$ and background outside the sphere. \textbf{(B)} Our proposed occlusion-aware paradigm parameterizes the scene into three sequential parts. The occlusion is explicitly modeled within the interval space between the camera and an inner sphere $\pi$ with a radius $\mathbf{r}<\mathbf{R}$.}
    
    \label{fig:scene_parameterization}
\end{figure}

To decouple the human from the background, the scene is usually parameterized separately \cite{guo2023vid2avatar}. This kind of parameterization first introduced by \cite{zhang2020nerf++} uses a sphere $\Pi$ to cover all of the space intended to be occupied by the human. The space outside the sphere is parameterized as the background by inverting the sphere (\figureautorefname~\ref{fig:scene_parameterization} (A)). Rendering is then achieved via ray composition. 

\noindent
\textbf{Motivation.} The above \textcolor{human_color}{human}$\rightarrow$\textcolor{background_color}{background} parameterization successfully renders the human and background when the human can be clearly viewed by the camera with no obstacles occluding the human. However, this ideal setting is impractical for in-the-wild videos, which may contain unexpected foreground objects other than the human. The object occlusions can interfere with human modeling. Considering this, we propose to parameterize the scene into three parts: \textcolor{obstacle_color}{occlusion}$\rightarrow$\textcolor{human_color}{human}$\rightarrow$\textcolor{background_color}{background} (\figureautorefname~\ref{fig:scene_parameterization} (B)).

\noindent
\textbf{\textcolor{human_color}{Human}.} For each frame in the video, we transform the deformed humans into the common canonical space, which is bounded by a sphere $\Pi$ with pre-defined radius $\mathbf{R}$. We learn a neural radiance field to model the surface of the human body implicitly (\sectionautorefname~\ref{sec:background}). We follow \cite{yariv2021volume,guo2023vid2avatar}'s approach to SDF-based volume rendering. For each point $\mathbf{x}^i$ in the canonical space, we first compute its signed distance $\mathbf{s}^i$ and then transform signed distances into density values $\sigma^i$ by applying the cumulative distribution function (CDF) of the scaled Laplace distribution to the negative of the signed distances. Then, we aggregate the radiance and density of each sample along the rays for volume rendering \cite{lombardi2019neural}:
\begin{equation}
     \mathbf{C} = \sum_{i}\mathbf{c}^i(1-\exp(-\mathbf{\sigma}^i\delta^i))\prod_{j<i}\exp(-\mathbf{\sigma}^j\delta^j),
\end{equation} \label{eq:volume_rendering}
\!
\begin{equation}
     \tau^i = \exp \left( -\sum_{j<i} \sigma^j \left(1 - \exp(-\sigma^j) \right) \right),
\end{equation} \label{eq:transmittance}
where $\delta^i$ is the z-axis distance between two ray samples and $\tau^i$ is the transmittance that indicates the probability of the ray not hitting anything within the interval $[1, i]$. To eliminate ambiguity when parameterizing different parts of the scene, we normalize the center of the human to $\mathbf{O}=\vec{0}$.


\noindent
\textbf{\textcolor{background_color}{Background}.} 
\textcolor{Highlight}{To handle the unbounded environment outside a scene’s foreground, we adopt the \emph{inverted-sphere} scheme from NeRF++ \cite{zhang2020nerf++}. We treat any point whose distance from the origin exceeds the radius $\mathbf{R}$ of an outer sphere $\Pi$ as background. Such a point \(\mathbf{x}=[x,y,z]\) is represented as the 4-tuple \(\mathbf{x}_{\text{bg}}=[x_{\text{bg}},y_{\text{bg}},z_{\text{bg}},1/d]\): the first three entries store its direction on the outer sphere surface, while \(1/d\) (inverse distance) compresses the infinite radial range into the bounded interval \((0,1]\). Sampling uniformly in \(1/d\) concentrates rays near the foreground, preserves coverage of far-field content, and keeps all coordinates numerically stable \cite{zhang2020nerf++}.} Similar to \cite{guo2023vid2avatar}, the rendering of the background is conditioned on the combination of a per-frame latent code that is constantly optimized during training, 4D parameterization, and embedded ray directions.

\noindent
\textbf{\textcolor{obstacle_color}{Occlusion}.} On top of the two-part scene parameterization, in this work, we propose to cater for obstacles that come between the human and the camera when filming the video. We define the occlusion to lie inside the sphere $\Pi$ and occupy a bounded sub-space within $\Pi$. To parameterize the human and occlusion separately, a concentric inner sphere $\pi$ is introduced with radius $\mathbf{r}<\mathbf{R}$. Without any semantic priors about the obstacles (e.g. size, appearance), we use the \emph{entire space} between the camera and $\pi$ to learn the neural radiance field for occlusion. To fully bound this sub-space, $\pi$ is built as an inscribed sphere to the outermost rays shot from the camera, and its radius $\mathbf{r}$ is determined accordingly to ensure every ray intersects $\pi$ at least once. For any radius $r$, the number of intersections can be determined via $(\mathbf{o}\cdot\mathbf{d})^2 - ||\mathbf{o}^2|| + r^2$, where $\mathbf{o}$ is the camera location, $\mathbf{d}$ is the ray direction, and $||\cdot||$ is L1 norm on the coordinate dimension. There is no intersection between the rays and $\pi$ when the above function is negative, and there must be at least one intersection otherwise. Given the monotonicity of the function w.r.t $r$, we can easily find the minimum possible $\mathbf{r}$ through binary search within $(0, \mathbf{R})$ during pre-processing.


Since we do not focus on rendering quality for the obstacles and background, we improve network efficiency by using the same rendering network $\mathcal{F}^{Scene}$ for both obstacles and background while conditioning on different per-frame latent codes. For the unbounded background, the quadruple $\mathbf{x}_{bg}$ must be computed through vector rotation which is time-consuming. In the bounded occlusion space, the quadruple $\mathbf{x}_{occ}$ can be computed more efficiently by solving the following quadratic equation with respect to $t$:
\begin{equation}
\begin{gathered}
||\mathbf{o}+t\mathbf{\mathbf{d}}|| = \mathbf{r}, \\
||\mathbf{d}||^2t^2 + 2(\mathbf{o}\cdot \mathbf{d})t + ||\mathbf{o}||^2 - \mathbf{r} = \vec{0},\\
 t = \frac{-\sqrt{||\mathbf{\mathbf{o}} \cdot \mathbf{d}||^2 + \left( ||\mathbf{r}^2|| - ||\mathbf{\mathbf{o}}^2|| \right) ||\mathbf{d}^2||} - ||\mathbf{\mathbf{o}} \cdot \mathbf{d}||}{||\mathbf{d}^2||}.
\end{gathered}
\end{equation}
The parameterized 3D coordinates on the surface of $\pi$ can be then computed as $\{x_{occ},y_{occ}, z_{occ}\}=\mathbf{o} + t\mathbf{d}$.

Sharing the same network $\mathcal{F}^{scene}$ for both background and occlusion requires the inputs to be at the same magnification scale. To do this, we normalize both $\{x_{occ}, y_{occ}, z_{occ}\}$ and $\{x_{bg}, y_{bg}, z_{bg}\}$ to the unit sphere (\figureautorefname~\ref{fig:scene_parameterization} (C)). Such normalization will not lead to overlapped surface vectors since $\mathbf{r}$ is strictly smaller than $\mathbf{R}$. To further decouple the dependencies for occlusion/background rendering, we determine the quadruple $\mathbf{x}_{occ}$ as the concatenation of $\{x_{occ},y_{occ}, z_{occ}\}$ and \emph{negation} of the inverted depth $-\frac{\mathbf{r}}{||\mathbf{x}^2||}$.

\begin{figure}[t]
    \centering
    \includegraphics[width=0.99\linewidth]{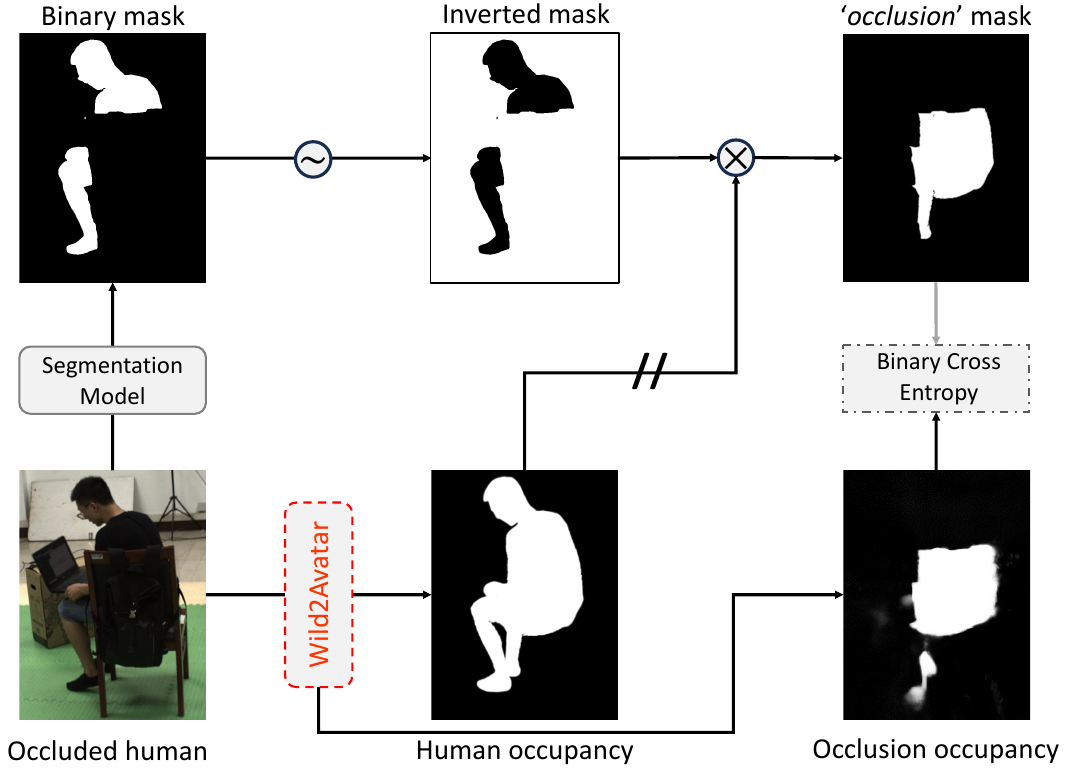}
    \caption{\textbf{Workflow for the occlusion decoupling loss $\mathcal{L}_{occ}$.} `$\sim$' indicates binary inversion and `//' indicates gradient stopping.}
    \label{fig:ob_loss}
\end{figure}

\noindent
\textbf{Scene Composition.} We cast different sets of ray samples on the neural fields individually and query their color via \equationautorefname~\ref{eq:volume_rendering} separately. For the bounded occlusion and human, we sample $\mathbf{x}_{occ}$ and $\mathbf{x}_{fg}$ directly. For the background, we sample $\frac{1}{d}$ instead and compute $\mathbf{x}_{bg}$ afterwards. The ray opacity $\alpha$ for each neural field can be determined via $\alpha=\sum_{i} \tau^i$ (\equationautorefname~\ref{eq:transmittance}). The topology of the scene is explicitly constrained through sequential composition:
\begin{equation}
    \mathbf{C} = \mathbf{C}_{occ} + (1 - \alpha_{occ})\mathbf{C}_{fg} +  (1 - \alpha_{occ})(1 - \alpha_{fg})\mathbf{C}_{bg}.
\end{equation}






\subsection{Optimization Objectives} \label{sec:objectives}

\noindent
\textbf{Rendering objectives.}
The most common objective for neural rendering models is the pixel-wise photometric loss $\mathcal{L}_{rgb}$ \cite{nerf} that forces renderings to reconstruct input images. For SDF-based rendering, a regularizer $\mathcal{L}_{eik}$ \cite{gropp2020implicit} is usually used to constrain the implicit geometry to satisfy the Eikonal equation. Moreover, following the two-part scene decomposition paradigm \cite{zhang2020nerf++}, we borrow the decomposition loss $\mathcal{L}_{dec}$ \cite{guo2023vid2avatar} for better decoupling of the dynamic human from the static background.

\begin{figure*}[!t]
    \centering
    \includegraphics[width=0.85\linewidth]{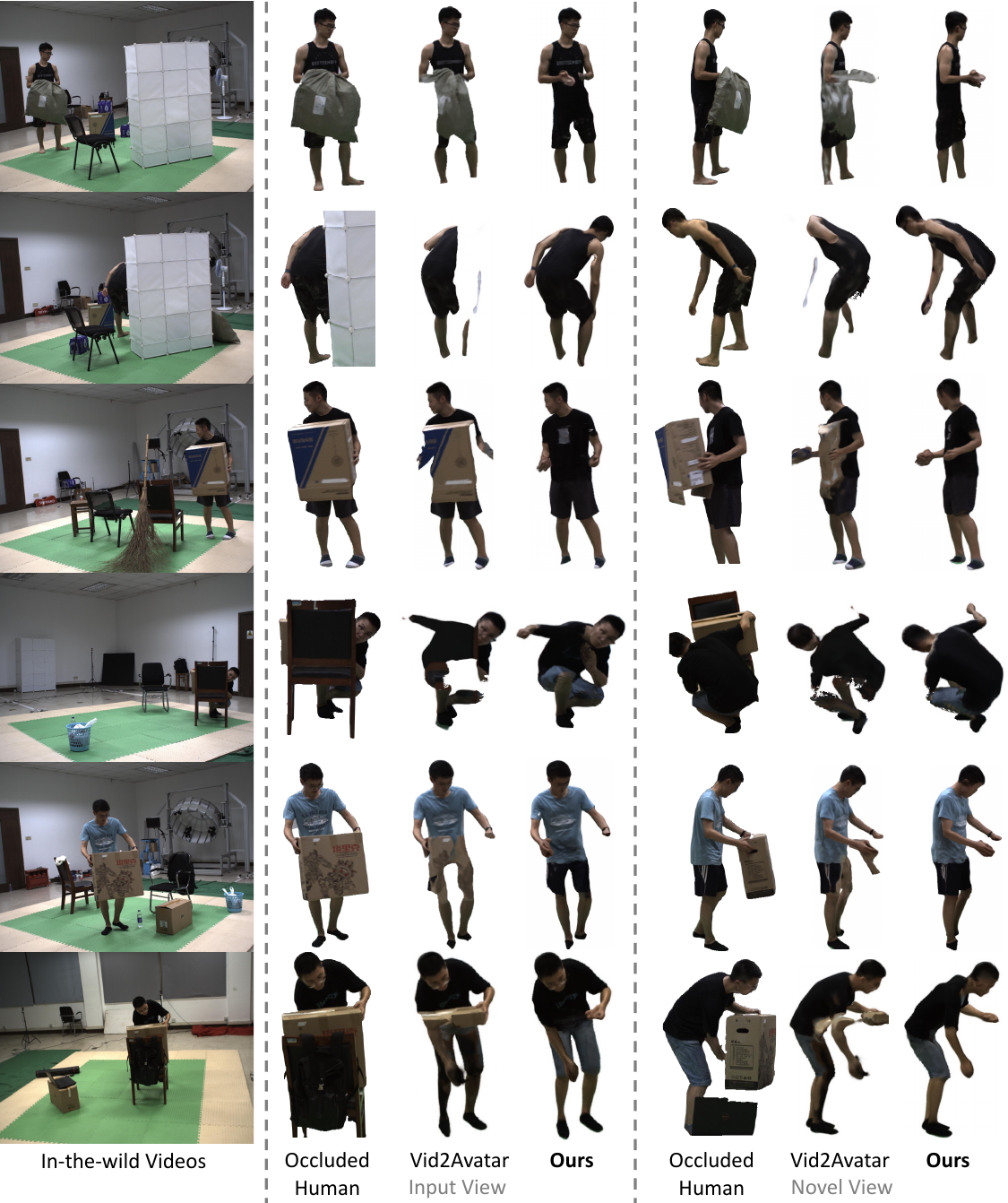}
    \caption{Qualitative comparisons with Vid2Avatar \cite{guo2023vid2avatar} on 6 occluded sequences in the OcMotion dataset \cite{huang2022occluded}.}
    \label{fig:ocmotion_result}
\end{figure*}

\noindent
\textbf{Occlusion decoupling.} In videos with occlusions, optimizing towards the above objectives tends to merge obstacles with the human and mistakenly render obstacle textures as part of the human appearance (\figureautorefname~\ref{fig:ocmotion_result}). So, we propose an additional objective that encourages the decoupling of the occluding obstacles and the human. Without any external knowledge of the obstacles occluding the human, it is impossible to model them completely. Hence we only parameterize potential obstacles within the interval between the camera and the human. To be more specific, we encourage the obstacles' density to be high at the pixels of human occupancy but are invisible from the camera. The visibility of the human can be determined via a binary mask $\mathbf{M}$ from an off-the-shelf segmentation model and the occupancy of the human can be obtained from volume rendering:
\begin{equation}
    \mathcal{L}_{occ} = \texttt{BCE}\left(\alpha_{occ}, (1- \mathbf{M})(\texttt{sg}(\alpha_{fg}) > \epsilon)\right),
\end{equation}
where pixels in $\alpha_{occ}, \alpha_{fg}\in[0,1]$ are the obstacle and human densities respectively, $\epsilon=0.1$ is a threshold to determine the occupancy of the human, $\texttt{BCE}(\cdot)$ is the binary cross entropy function, and $\texttt{sg}(\cdot)$ indicates gradient stopping. The workflow is outlined in \figureautorefname~\ref{fig:ob_loss}. We noticed that occlusions only appear in a small area of most videos, and the remaining areas usually have empty obstacle densities. To deal with this data imbalance, we use weighted $\texttt{BCE}(\cdot)$ and assign a higher weight to the pixels that tend to be occluded.

\noindent
\textbf{Geometry completeness regularization.} In the decomposition loss $\mathcal{L}_{dec}$, rays that intersect the surface of the human are encouraged to have high occupancy of the 2D space. We observed that such regularization in 2D introduces ambiguity in 3D and the rendered human geometry is usually incomplete under strong occlusions. Considering this, we propose to regularize the completeness of 3D geometry by enforcing the ray samples near the surface $\mathcal{X}_{near}$ to be closer to the implicit surface:
\begin{equation}
    \mathcal{L}_{comp} = \frac{1}{|\mathcal{X}_{near}|}\sum_{x\in \mathcal{X}_{near}}|\mathbf{s}(x)|,
\end{equation}
where $\mathcal{X}_{near}$ can be determined by the Euclidean distances to the human mesh, which was initialized as SMPL \cite{SMPL} mesh and updated during optimization \cite{guo2023vid2avatar}.

\methodname\ is trained to optimize the combination of all of the above objectives, each with a scale weight $\lambda_{(\cdot)}$:
\begin{equation}
    \mathcal{L}_{rgb} + \lambda_{eik}\mathcal{L}_{eik} + \lambda_{dec}\mathcal{L}_{dec} + \lambda_{occ}\mathcal{L}_{occ} + \lambda_{comp}\mathcal{L}_{comp}.
\end{equation}




\section{Experiments}
\label{sec:Experiments}

\begin{figure*}[t]
    \centering
    \includegraphics[width=0.85\linewidth]{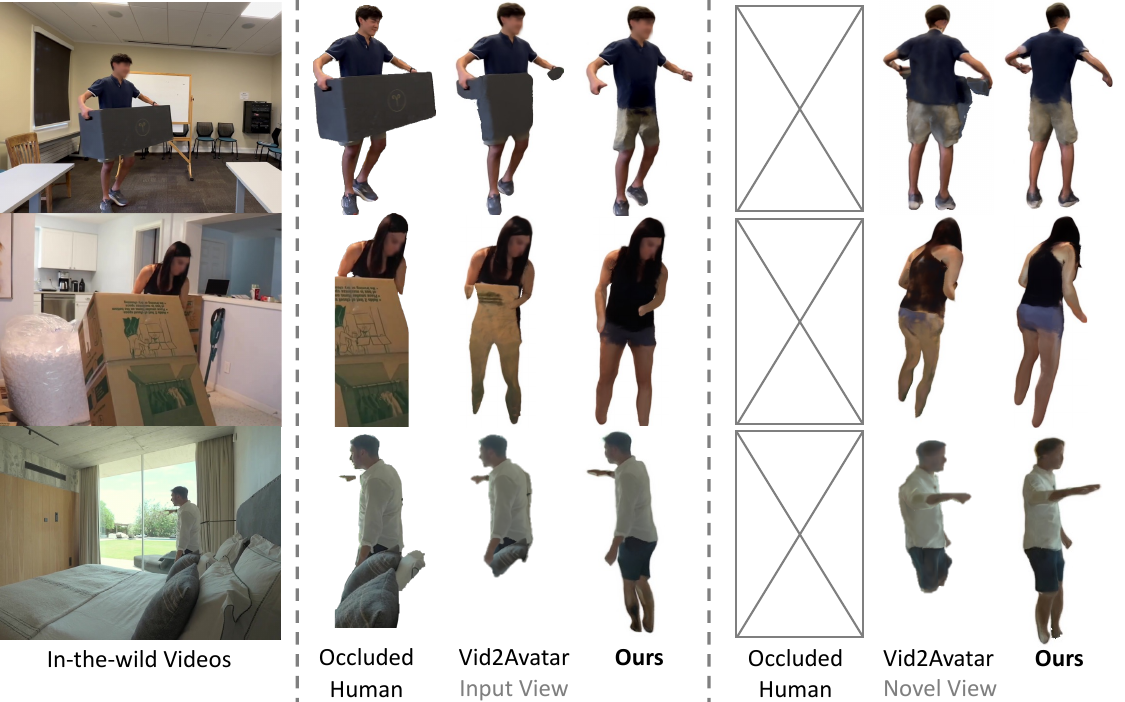}
    \caption{Comparisons with Vid2Avatar \cite{guo2023vid2avatar} on 3 occluded in-the-wild monocular videos. Occluded humans from novel views are not available. We manually blurred the faces for the in-the-wild videos to maintain anonymity.}
    \label{fig:inthewild_result}
\end{figure*}

\begin{figure*}[t]
    \centering
    \includegraphics[width=0.85\linewidth]{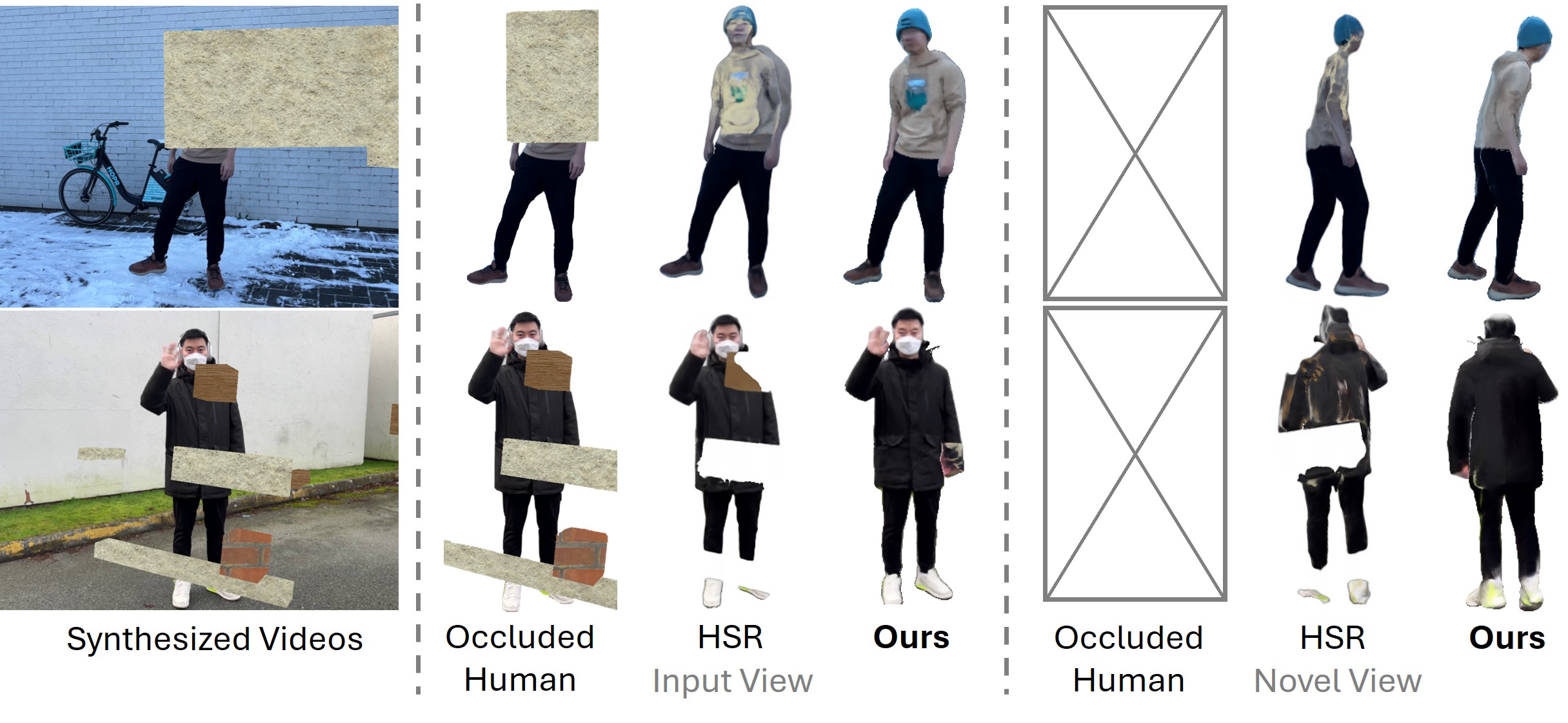}
    \caption{\textcolor{Highlight}{Comparisons with HSR \cite{xue2024hsr} on 2 synthesized occluded videos. Occluded humans from novel views are not available.}}
    \label{fig:quantitative_results_synthesis}
\end{figure*}

\subsection{Datasets}
Unlike our prior work that was majorly evaluated on simulated occlusions \cite{xiang2023rendering}, we focus more on evaluations on real-world occlusions in this extension work. 

\noindent\textbf{OcMotion \cite{huang2022occluded}.} This dataset consists of indoor scenes of humans engaging with a variety of objects while being partially occluded by them. We used 6 challenging sequences out of the 48 videos at different levels of occlusion to evaluate the methods. In particular, we drew only 100 frames from each of the videos to train the models. We initialized the optimization process with the dataset's provided camera matrices, human poses, and SMPL parameters. Frame-wise binary human segmentation masks are obtained from the Segment Anything Model \cite{kirillov2023segment}.

\noindent\textbf{In-the-wild videos.} We conducted additional experiments on 3 real-world videos, two of which was downloaded from YouTube with CC license while the other one was captured by our research team using a cell phone camera. We sampled $~$150 frames from all of the videos for training. In these videos, we obtained camera matrices, human poses, and SMPL parameters using SLAHMR \cite{goel2023humans,ye2023slahmr}. Since no ground truth poses are provided, evaluations on these videos also indicate the robustness of methods to inaccurately estimated priors.



\noindent\textcolor{Highlight}{\textbf{Synthesized videos.} To better evaluate our model under challenging conditions such as moving cameras and long-horizon occlusions, we synthesized occluded variants of 5 of the NeuMan~\cite{jiang2022neuman} videos by inserting textured, box-shaped static occluders into each scene, guided by the 3D human trajectory and camera parameters from COLMAP~\cite{schoenberger2016sfm}. These occluders are 3D-consistent with the background and rendered directly into the frames. Additional experiments were conducted on the 5 synthesized videos, bringing the total to 14—representing the most extensive evaluation of occluded human modeling with moving cameras to date. Similar to in-the-wild videos, we obtained SMPL parameters using SLAHMR \cite{ye2023slahmr}.}

\textcolor{Highlight}{Due to the scarcity of occluded human videos captured by moving cameras, most of our evaluations were conducted on videos recorded with stationary cameras, except for one in-the-wild video, whose result is shown in the bottom row of Figure~\ref{fig:inthewild_result}. However, as an extension of Vid2Avatar~\cite{guo2023vid2avatar}, our method is not limited to fixed-camera scenarios and can be naturally applied to moving-camera setups.}

\subsection{Implementation Details} Following the common NeRF protocol, we optimize human NeRFs on a per-video basis. Our $\mathcal{F}^{fg}$ consists of two sub-networks: one for learning the implicit SDF (implicit network) and one for learning the point-wise radiance (rendering network). The implicit network uses a MLP with 8 layers of 256 feature dimensions and the rendering network uses a MLP with 4 layers of 256 feature dimensions. Similar to $\mathcal{F}^{fg}$, $\mathcal{F}^{scene}$ also has two sub-networks but in different configurations: the implicit network uses a MLP with 8 layers of 256 feature dimensions and the rendering network uses a single-layer MLP  of 256 feature dimension. Both \textcolor{background_color}{background} and \textcolor{obstacle_color}{occlusion} learn per-frame latent codes of 32 feature dimensions. We use only 64 ray samples for training and 128 ray samples for evaluation/inference \cite{guo2023vid2avatar}. During training, 512 rays are randomly sampled within the 2D human bounding box. Note that we use the same binary human segmentation mask for both Vid2Avatar and \methodname\ for fairness. The loss weights are set as $\lambda_{eik}=0.1$, $\lambda_{dec}=0.003$, $\lambda_{comp.}=0.2$, and $\lambda_{occ}=0.1$. We adopted the Adam \cite{kingma2014adam} optimizer with an initial learning rate of $5.0e^{-4}$, which is decayed by $\frac{1}{2}$ at the 200$_{th}$ and the 500$_{th}$ step.


\begin{table*}[t]
    \centering

    \begin{tabular}{l|ccc|ccc}
    \toprule
    & \multicolumn{3}{c|}{OcMotion} & \multicolumn{3}{c}{In-the-wild} \\
    Methods & $\texttt{quality}_{vis}$ & $\texttt{comp.}$ & $\texttt{quality}_{user}$ & $\texttt{quality}_{vis}$ & $\texttt{comp.}$ & $\texttt{quality}_{user}$ \\  
    \hline
     Vid2Avatar \cite{guo2023vid2avatar} & 12.74 & 0.74& 3.7$\pm$1.3  & 8.84 & 0.66 & 3.6$\pm$1.0\\
      OccNeRF$_{500}$ \cite{xiang2023rendering} & 12.33 & 0.76 & 4.0$\pm$0.9 & 7.98 & 0.70 & 3.2$\pm$1.1\\
     Wild2Avatar &   \cellcolor{best_color}13.01 &  \cellcolor{best_color}0.81 &  \cellcolor{best_color}6.6$\pm$0.3  &  \cellcolor{best_color}9.68  &  \cellcolor{best_color}0.74 &  \cellcolor{best_color}7.0$\pm$0.8\\
    \bottomrule
    \end{tabular}

    \caption{Quantitative comparisons on the three datasets. We color cells that have the \colorbox{best_color}{best} metric values.}
    \label{tab:quantitative_results}
\end{table*}

\begin{table}[t]
    \centering

    \begin{tabular}{l|ccc}
    \toprule
     Methods & PSNR$\uparrow$ & SSIM$\uparrow$ & LPIPS$\downarrow$\\  
    \hline
     HSR \cite{xue2024hsr} & 18.77 & 0.9005 & 68.80\\
     Wild2Avatar & \cellcolor{best_color}22.47 & \cellcolor{best_color}0.9624 & \cellcolor{best_color}44.36\\
    \bottomrule
    \end{tabular}

    \caption{\textcolor{Highlight}{Quantitative comparisons on the synthesized videos. We color cells that have the \colorbox{best_color}{best} metric values.}}
    \label{tab:quantitative_results_synthesis}
\end{table}


\subsection{Evaluations} \label{sec:evaluations}
\begin{figure}[t]
    \centering
    \includegraphics[width=0.95\linewidth]{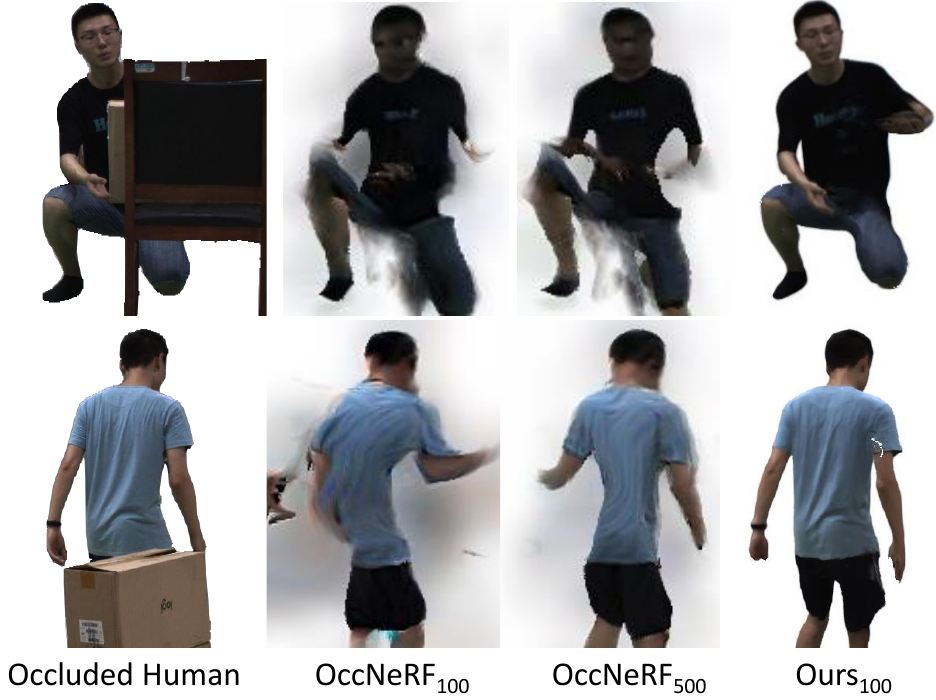}
    \caption{Comparison against OccNeRF \cite{xiang2023rendering} trained on 100 and 500 training frames.}
    \label{fig:occnerf_result}
\end{figure}

\noindent\textbf{Comparisons.} \methodname\ is mainly compared against Vid2Avatar \cite{guo2023vid2avatar}, which is the state-of-the-art in human rendering. We also compare against the previous occlusion-aware human modeling counterpart - OccNeRF \cite{xiang2023rendering}. For fairness, all comparison methods use the same binary human segmentation masks either for supervision or ray sampling. Note that OccNeRF originally requires more than 500 frames for stable training while \methodname\ and Vid2Avatar only require $\sim$100. We conducted experiments on both high and low frame count regimes for OccNeRF in \figureautorefname~\ref{fig:occnerf_result}. 

\textcolor{Highlight}{NeuMan~\cite{jiang2022neuman}, HOSNeRF~\cite{liu2023hosnerf}, and HSR~\cite{xue2024hsr} are similar methods that also use layered representations of the scene. However, they have significant limitations preventing us from fully utilizing them as baselines. NeuMan learns a scene NeRF model, which is not feasible when the camera is stationary. On the other hand, the official HOSNeRF implementation cannot extend to custom videos beyond the six videos presented in their paper, and fails when the occlusion is not connected to the human. Additionally, both NeuMan and HOSNeRF require more resources than our method: NeuMan requires 3 GPUs for a total of 175 hours of training time, HOSNeRF requires 4 GPUs for a total of 118 hours of training time, while \methodname~requires only a single 24GB GPU for 48 hours of training time. To better evaluate Wild2Avatar, we provide comparison results against HSR on 2 of the synthesized videos, which are shown in \figureautorefname~\ref{fig:quantitative_results_synthesis} and quantified in \tableautorefname~\ref{tab:quantitative_results_synthesis}.}

\noindent\textbf{Metrics.} We used qualitative and quantitative evaluations to compare the methods. For qualitative evaluations, we render the human behind the obstacles as well as novel views to assess the capability of occlusion handling. Furthermore, we rely on three quantitative metrics to measure the capability of renderers from different perspectives. First, we compute the commonly used Peak Signal-to-Noise Ratio (PSNR) on the visible human parts (determined by the binary segmentation mask) to measure the rendering quality on unoccluded human appearance. Then, we calculate the Intersection over Union (IoU) between human occupancy masks and  GT SMPL mesh silhouettes to measure the completeness of rendering \cite{xiang2023rendering}. Lastly, for a more rational assessment of the rendering quality of occluded humans, we conducted user studies to ask 25 out-of-field people to rate human renderings at a scale of 0-10 (the higher the better). The three metrics are denoted as $\texttt{quality}_{vis}$, $\texttt{comp.}$, and $\texttt{quality}_{user}$ respectively.

\textcolor{Highlight}{For evaluations on the synthesized videos, we follow the common protocol and calculated PSNR and SSIM values between humans in the unoccluded raw videos and the rendered humans at the input views. }

\begin{figure*}[h]
    \centering
    \includegraphics[width=0.95\linewidth]{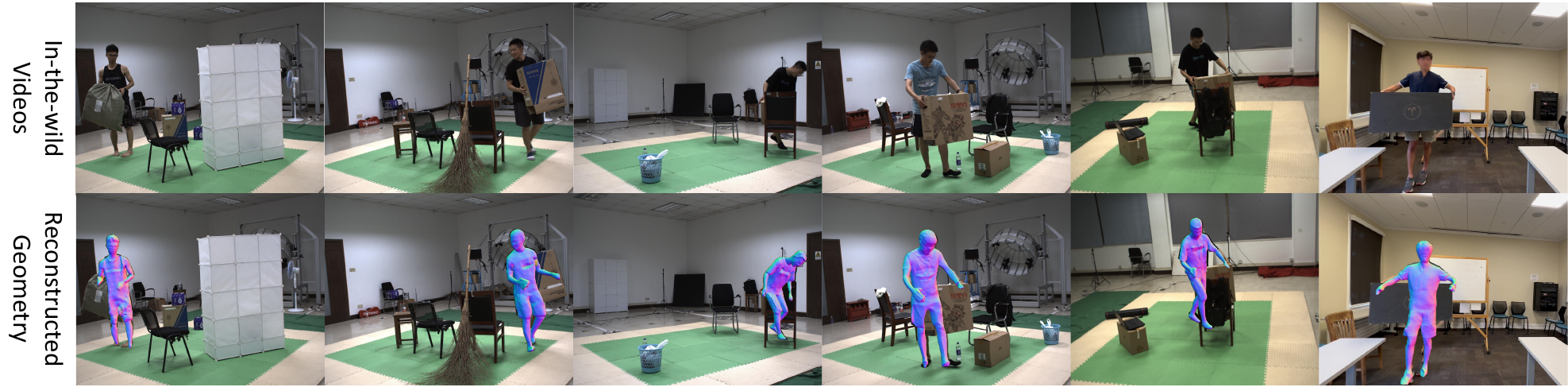}
    \caption{Geometry reconstructions. We manually blurred the faces for the in-the-wild videos for maintaining anonymity.}
    \label{fig:geo_recon}
\end{figure*}

\begin{figure*}[t]
    \centering
    \includegraphics[width=0.95\linewidth]{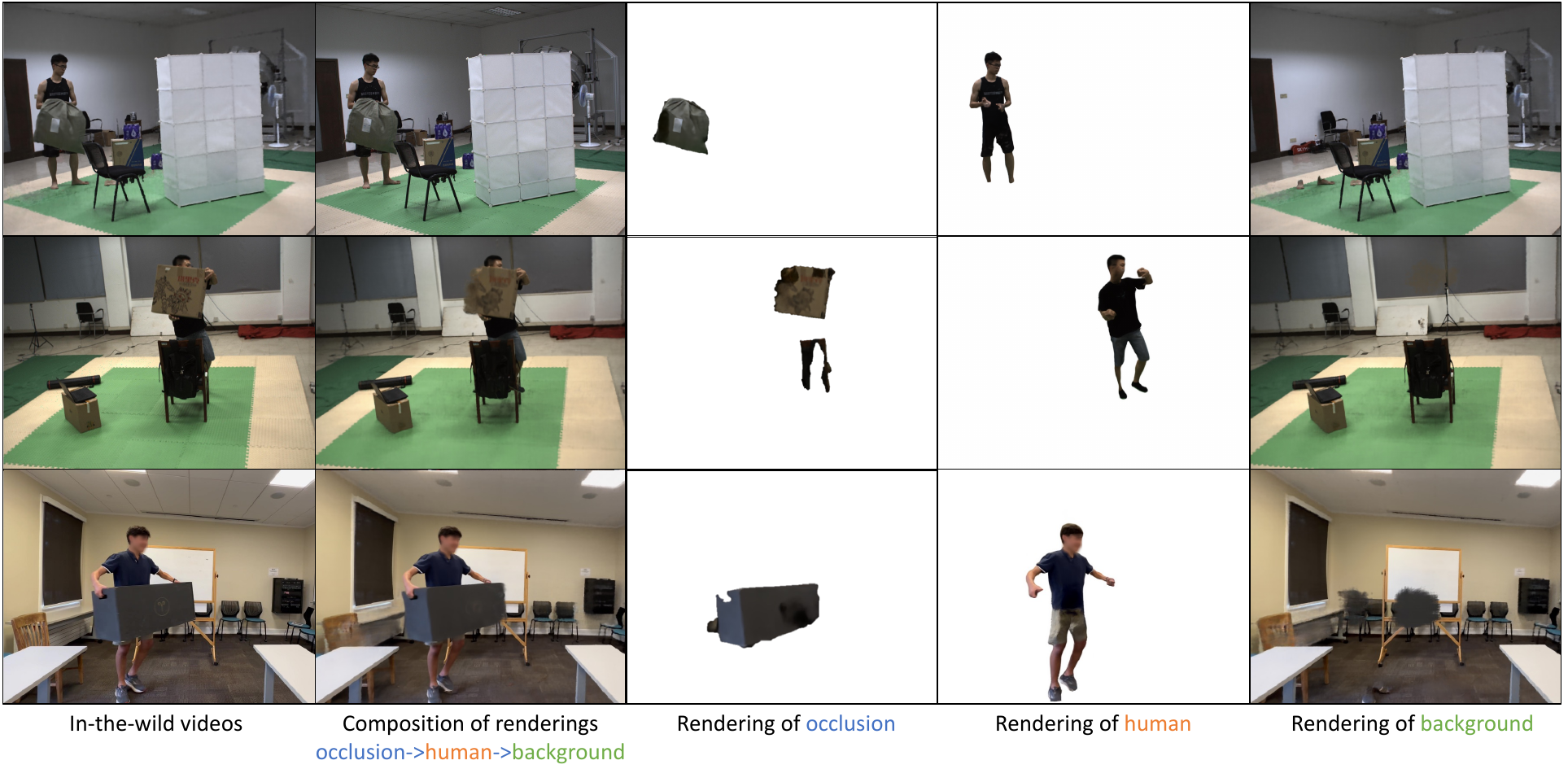}
    \caption{Visualizations of scene decomposition. \methodname\ separately renders \textcolor{obstacle_color}{occlusion}, \textcolor{human_color}{human}, and \textcolor{background_color}{background}.}
    \label{fig:scene_decomposition}
\end{figure*}

\subsection{Results}
We first compare the results of Vid2Avatar and \methodname\ on the OcMotion videos in \figureautorefname~\ref{fig:ocmotion_result}, where both methods demonstrate high fidelity in reconstructing the visible human parts without obvious artifacts or floaters, showcasing the superiority of representing humans via implicit SDF. However, Vid2Avatar struggles with recovering the occluded body parts and erroneously renders parts of the occlusions as human appearance. \methodname\ effectively decouples occlusions from the human body and recovers consistent human appearances. \figureautorefname~\ref{fig:inthewild_result} shows results on more challenging in-the-wild videos. Wild2Avatar consistently outperforms Vid2Avatar in real-world scenarios.

Subsequently, we extend our comparison to include \methodname\ against the recent occlusion-aware human modeling method OccNeRF \cite{xiang2023rendering}, as illustrated in \figureautorefname~\ref{fig:occnerf_result}. For a fair comparison, OccNeRF was trained on datasets of 500 frames and 100 frames, respectively. Despite its ability to recover occluded human parts, OccNeRF, is prone to common defects such as floaters and artifacts, often resulting in the body appearing unexpectedly twisted and thus leading to a lower rendering quality.

In \tableautorefname~\ref{tab:quantitative_results}, the quantitative results reveal on-par rendering performances for visible parts, with \methodname\ consistently outperforming Vid2Avatar in terms of body geometry and the rendering quality of occluded parts.

\textcolor{Highlight}{Moreover, we include comparisons between \methodname\ and HSR~\cite{xue2024hsr} - the most recent method that also decouples occlusions from human subjects - on the synthesized videos. As shown in \figureautorefname~\ref{fig:quantitative_results_synthesis} and \tableautorefname~\ref{tab:quantitative_results_synthesis}, our rendering results exhibit significantly higher fidelity, with fewer occlusion artifacts and more complete human bodies. These results confirm that \methodname\ not only performs well on videos with moving cameras but also outperforms the current state-of-the-art.}

\subsection{Geometry Reconstructions}

When rendering the human avatars through volumetric rendering, we can not only obtain their appearances but also reconstruct their underlying geometries. We overlay the reconstructed human surface normals on the input frames in \figureautorefname~\ref{fig:geo_recon}. The results show that \methodname\ excels in reconstructing human geometries from occluded videos.

\subsection{Ablation Studies}

\begin{figure*}[t]
    \centering
    \includegraphics[width=0.95\linewidth]{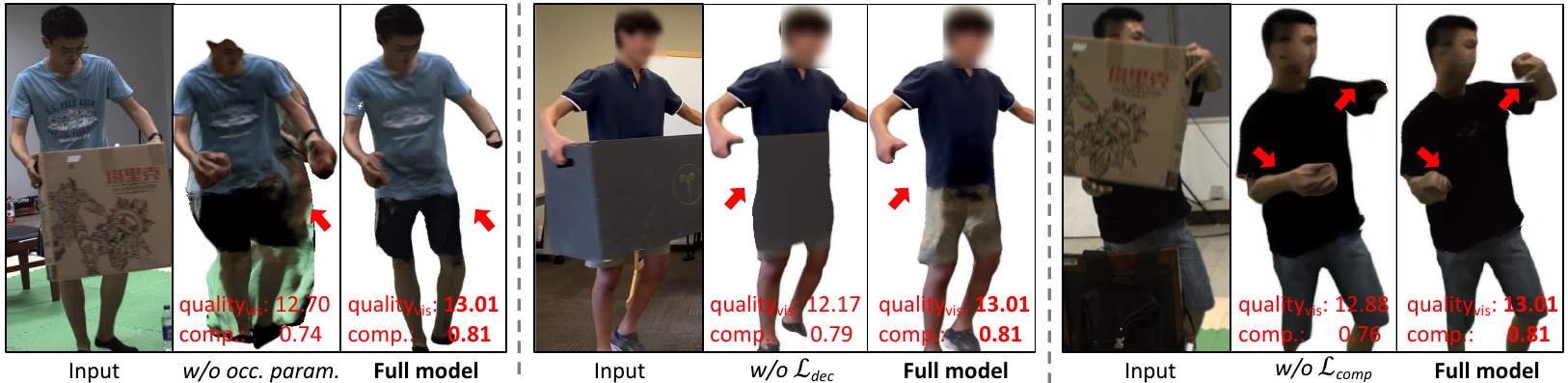}
    \caption{Ablation results. Major differences are highlighted.}
    \label{fig:ablations}
\end{figure*}

\noindent\textbf{Impact of occlusion-aware parameterization.}  One of the major contributions that enables \methodname\ to achieve human renderings under occlusions is the occlusion-aware scene parameterization. Here we verify the effectiveness of our design by replacing it with an alternative optimization strategy that simply optimizes human NeRF on the visible parts only \cite{xiang2023rendering} while maintaining all the objectives except $\mathcal{L}_{occ}$. As shown in the first column of \figureautorefname~\ref{fig:ablations}, even though \methodname\ can still recover the occluded appearances without the proposed parameterization, the rendered results are corrupted with many artifacts due to data scarcity.

\noindent\textbf{Impact of $\mathcal{L}_{occ}$.}  In \sectionautorefname~\ref{sec:objectives}, the occlusion decoupling loss is proposed to further disentangle the rendering of human and occlusions. We validated the effect of this loss by simply removing it from optimization objectives. As can be seen in the second column of \figureautorefname~\ref{fig:ablations}, the proposed loss is necessary to fully recover the occluded region of the human.

\noindent\textbf{Impact of $\mathcal{L}_{comp}$.}  As noted in \cite{xiang2023rendering}, the completeness of the 3D human geometry needs to be explicitly enforced during training. When removing the completeness loss, the 2D renderings can easily degenerate. We also observed that the proposed loss serves as a regularizer to force the human geometry to be consistent with the SMPL mesh, which helps prevent the model from rendering incorrect poses (third column in \figureautorefname~\ref{fig:ablations}).

\noindent\textcolor{Highlight}{\textbf{Impact of optimizing inaccurate SMPL jointly with Wild2Avatar.} For our evaluations on both in-the-wild and synthesized videos, SMPL parameters were estimated using an occlusion robust HMR model~\cite{zhu2024dpmesh}. As shown in \figureautorefname~\ref{fig:inthewild_result} and \figureautorefname~\ref{fig:quantitative_results_synthesis}, Wild2Avatar demonstrates strong generalization to inaccurate SMPL parameters. To further validate this, we conducted additional experiments on OcMotion videos, jointly optimizing predicted SMPL parameters and Wild2Avatar. As highlighted by the red arrows in \figureautorefname~\ref{fig:optimize_smpl}, this joint optimization progressively refines the SMPL poses, ultimately achieving reconstruction quality comparable to that obtained using ground-truth SMPL parameters.}

\subsection{Visualizations of Scene Decomposition}
As introduced in \sectionautorefname~\ref{sec:methods}, \methodname\ renders three scene parts compositionally. Human and background/occlusion are separately modeled in two different neural fields. Here we show individual renderings of the three scene parts in \figureautorefname~\ref{fig:scene_decomposition}. Note that since this work focuses solely on human rendering, artifact-free rendering of background and occlusions are outside the scope of this work. 

\subsection{Generative Ability}
\begin{figure*}[t]
    \centering
    \includegraphics[width=0.95\linewidth]{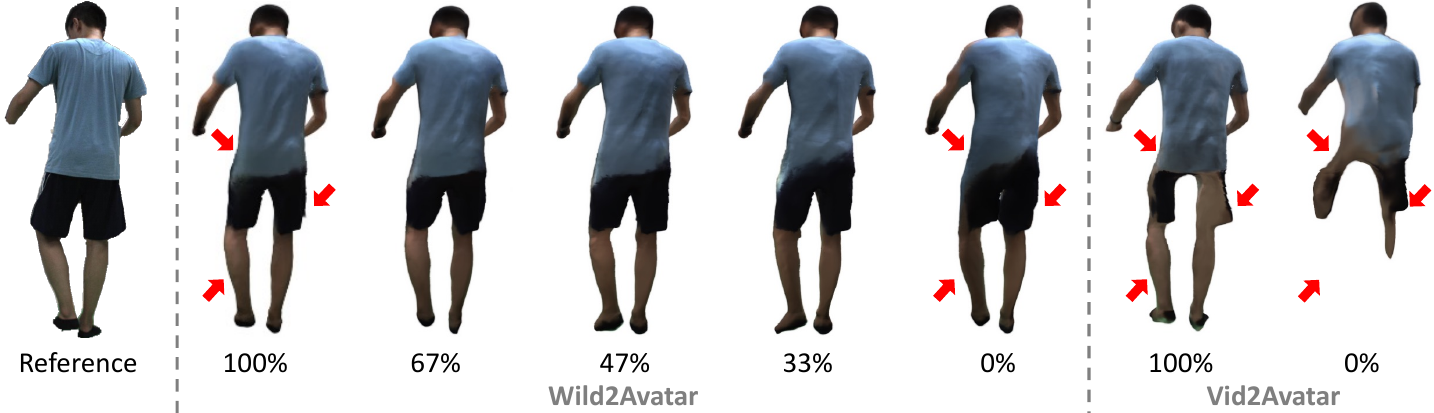}
    \caption{Study of generative ability. Qualitative comparison with Vid2Avatar \cite{guo2023vid2avatar} on amount of training images that show the full back of the human.}
    \label{fig:gen}
\end{figure*}

\begin{figure}[t]
    \centering
    \includegraphics[width=0.95\linewidth]{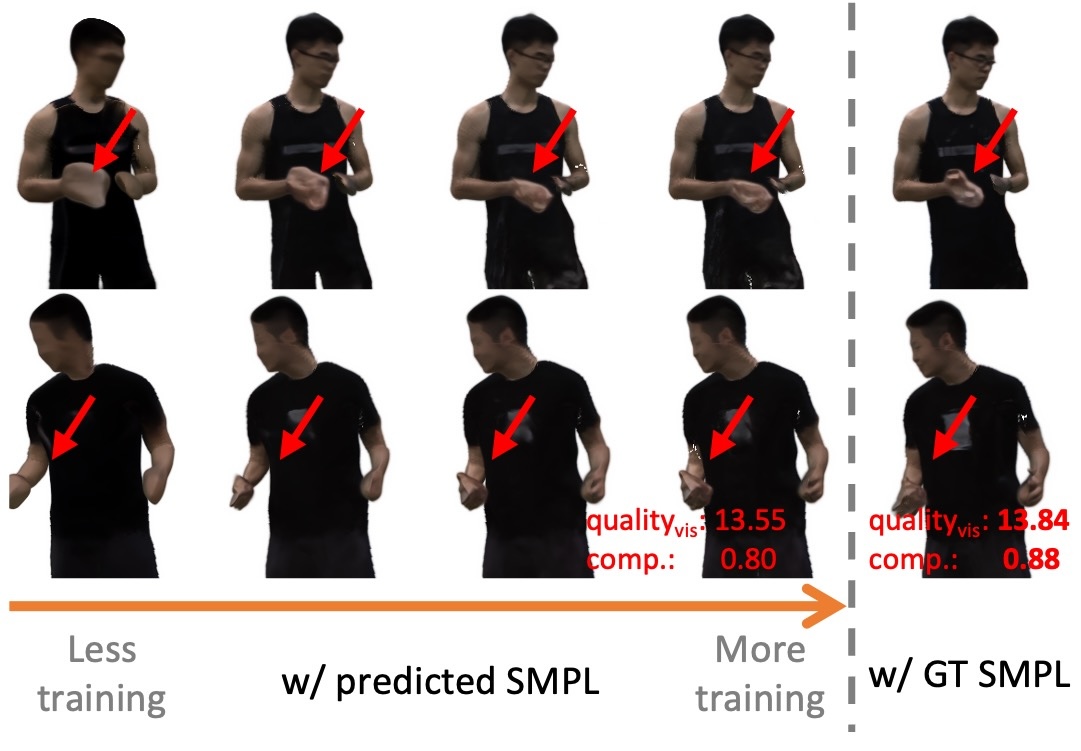}
    \caption{\textcolor{Highlight}{Comparison between joint training with predicted SMPL parameters (left) and GT parameters (right). Inaccurate limb rotations in the predicted SMPL can be progressively corrected through joint optimization.}}
    \label{fig:optimize_smpl}
\end{figure}

While existing methods to render occluded humans like OccNeRF \cite{xiang2023rendering} assume that every part of the human body needs to be visible at some point of the sequence, Wild2Avatar is able to hallucinate reasonable appearances for unobserved human parts. This is due to the design of the three-stage scene parameterization that enables standard photometric fitting on both visible and invisible human parts without forcing the model to overfit on the visible parts. We conducted supportive experiments to showcase Wild2Avatar's generative ability by gradually removing frames that show the back of the human. As shown in \figureautorefname~\ref{fig:gen}, Wild2Avatar still renders a complete and reasonable human geometry even without a clear view of the human's back.

\section{Discussion and Conclusion}
\label{sec:Discussion}
\label{sec:Conclusion}

\textcolor{Highlight}{\noindent\textbf{Discussion.} Training human NeRFs to fit input frames while handling occlusions is challenging. Without any semantic priors of the occluders, fully unsupervised occlusion-human decoupling is impractical. Our proposed occlusion decoupling loss $\mathcal{L}_{occ}$ provides a weakly-supervised solution using precomputed binary human segmentation masks. Both Vid2Avatar and \methodname\ rely on accurate pose estimates, which can lead to poor or incomplete rendering if human poses are initialized inaccurately. Moreover, modeling occlusions increases inference time, slowing down the optimization process. While these challenges persist, human pose estimation under occlusion is an actively evolving field, with new occlusion-robust methods such as PARE~\cite{kocabas2021pare}, 3DNBF~\cite{zhang20233d}, DPMesh~\cite{zhu2024dpmesh}, GLAMR~\cite{yuan2022glamr}, and SLAHMR~\cite{ye2023slahmr} continuing to improve robustness and temporal consistency. Our method builds on these advancements and offers a practical and effective solution for high-quality reconstruction of human geometry and appearance under occlusions. Future work may incorporate additional pose correction steps~\cite{ye2023slahmr} and efficient NeRF variants~\cite{jiang2022instantavatar, muller2022instant, kerbl20233d} or Gaussian splatting~\cite{kerbl20233d} for faster training.}

\noindent\textbf{Conclusion.} We introduced \methodname, a method to render a 3D human avatar from an occluded in-the-wild monocular video. Unlike prior work, our method was able to decouple the human body from both background and obstacle, allowing for a complete rendering without artifacts. We accomplished this through novel occlusion-aware scene parametrization: modeling the occlusion, human, and background separately. Additionally, we introduced training objectives to improve the quality of our renderings. Our work achieves high-fidelity state-of-the-art rendering of occluded humans on challenging in-the-wild videos.

\section*{Acknowledgment.} This work was partially funded by the NIH Grant R01AG089169 and P41EB027060, Panasonic Holdings Corporation, the Gordon and Betty Moore Foundation, the Jaswa Innovator Award, Stanford HAI, and Stanford Wu Tsai Human Performance Alliance. 

\newpage

\section{Biography}
 



\begin{IEEEbiographynophoto}{Tiange Xiang}
is a CS Ph.D. candidate at Stanford University. He is affiliated with Stanford AI Lab \& Stanford Vision and Learning Lab. He is advised by Prof. Fei-Fei Li, co-advised by Prof. Scott Delp and Prof. Ehsan Adeli. His research focuses on 3D computer vision \& AI for healthcare, which is partially supported by Stanford Human-centered AI fellowship. .
\end{IEEEbiographynophoto}

\begin{IEEEbiographynophoto}{Adam Sun}
is a senior and 1st-year master's student at Stanford University. He is interested in 3D computer vision for scene understanding and medical applications. Adam is a student researcher at Stanford Vision and Learning Lab (SVL), where he is advised by Professors Fei-Fei Li and Ehsan Adeli at the Partnership in AI-Assisted Care.
\end{IEEEbiographynophoto}

\begin{IEEEbiographynophoto}{Dr. Scott Delp} is the James H. Clark Professor at Stanford University, where he holds appointments in Bioengineering, Mechanical Engineering, and Orthopaedic Surgery. He is the founding chair of Stanford’s Bioengineering Department and directs major initiatives including the Wu Tsai Human Performance Alliance, the Restore Center, and the Mobilize Center. A pioneer in movement science and digital health, he developed widely-used tools like OpenSim and has authored over 325 research papers. 
\end{IEEEbiographynophoto}

\begin{IEEEbiographynophoto}{Dr. Kazuki Kozuka}
is a manager at Panasonic Corporation. He received his Ph.D degrees from Nagoya Institite of Technology in 2009. In 2009, he joined Advanced Technology Laboratories, Panasonic Corporation, Japan, where he was involved in research on neuroscience, image recognition. He was also a visiting scholar in Stanford University between 2016 and 2019. His research interests lie in visual understanding through machine learning, mainly for computer vision and natural language processing.
\end{IEEEbiographynophoto}

\begin{IEEEbiographynophoto}{Dr. Fei-Fei Li}
is the inaugural Sequoia Professor in the Computer Science Department at Stanford University, and Co-Director of Stanford’s Human-Centered AI Institute. She served as the Director of Stanford’s AI Lab from 2013 to 2018. And during her sabbatical from Stanford from January 2017 to September 2018, Dr. Li was Vice President at Google and served as Chief Scientist of AI/ML at Google Cloud. Since then she has served as a Board member or advisor in various public or private companies.
\end{IEEEbiographynophoto}

\begin{IEEEbiographynophoto}{Dr. Ehsan Adeli}
is an assistant professor at Stanford University School of Medicine, Department of Psychiatry and Behavioral Sciences, where he directs the Stanford Translational Artificial Intelligence (STAI) in Medicine and Mental Health Lab. He is also affiliated with the computer science department, Stanford Vision and Learning (SVL) lab. With a Ph.D. in computer science \& artificial intelligence and postgraduate training in biomedical imaging \& computational neuroscience, Ehsan is applying his expertise to solve critical problems in healthcare and neuroscience. To this end, his team focuses on developing innovative methods to enhance our understanding of digital humans within the 3D world.
\end{IEEEbiographynophoto}

\bibliographystyle{IEEEtran}
\bibliography{main}

\end{document}